\RequirePackage{fix-cm}
\documentclass[twocolumn]{svjour3}          
\smartqed  

\usepackage{algorithm} 
\usepackage{algpseudocode}
\usepackage{cite}
\usepackage[pdftex]{graphicx}
\usepackage{ragged2e}
\usepackage[tight,footnotesize]{subfigure}
\usepackage{rotating}
\usepackage{graphicx}
\usepackage{amsmath,amssymb} 
\usepackage{array}
\usepackage{multirow}
\usepackage{colortbl}
\usepackage{amsfonts}
\usepackage{pifont}
\usepackage{xspace}
\usepackage{etoolbox}
\usepackage{overpic}
\usepackage{color}
\usepackage{microtype}
\usepackage{pgfplots}
\usepackage{graphicx}
\usepackage{subcaption}
\newcommand{\figref}[1]{Fig. \ref{#1}}
\newcommand{\tabref}[1]{Tab. \ref{#1}}

\usepackage{soul, color, xcolor}

\newcommand{\orcid}[1]{\href{https://orcid.org/#1}{\includegraphics[width=10pt]{ocrid.png}}}

\def\etal{{\em et al}}
\usepackage{pgfplots}  
\usepackage{hyperref}
\hypersetup{breaklinks=true,citecolor=blue, colorlinks}

\graphicspath{{./Imgs/}}
\DeclareGraphicsExtensions{.pdf,.jpg,.png}

\usepackage{silence}
\journalname{Research Article}

\usepackage{pgfplots}
\pgfplotsset{compat=1.7}
\begin{document}

\title{Patch is Enough: Naturalistic Adversarial Patch against Vision-Language Pre-training Models}

\titlerunning{Short form of title}        

\author{Dehong Kong \and
  Siyuan Liang \and 
  Xiaopeng Zhu \and
  Yuansheng Zhong \and
  Wenqi Ren
}

\authorrunning{F. Author \etal} 

\institute{
The first Author (corresponding author) and fifth Author are with the Shenzhen Campus of Sun Yat-sen University. (Email: kongdh@mail2.sysu.edu.cn, renwq3@mail.sysu.edu.cn)
 \\
The second Author is with the National University of Singapore. (Email: pandaliang521@gmail.com)
 \\
The third and fourth authors are with Guangdong Testing Institute of Product Quality Supervision. (Email: zhuxp@gqi.org.cn,zhongys@gqi.org.cn)
}

\date{Received: date / Accepted: date}

\maketitle

\begin{abstract}
Visual language pre-training (VLP) models have demonstrated significant success across various domains, yet they remain vulnerable to adversarial attacks. Addressing these adversarial vulnerabilities is crucial for enhancing security in multimodal learning. Traditionally, adversarial methods targeting VLP models involve simultaneously perturbing images and text. However, this approach faces notable challenges: first, adversarial perturbations often fail to translate effectively into real-world scenarios; second, direct modifications to the text are conspicuously visible.
To overcome these limitations, we propose a novel strategy that exclusively employs image patches for attacks, thus preserving the integrity of the original text. Our method leverages prior knowledge from diffusion models to enhance the authenticity and naturalness of the perturbations. Moreover, to optimize patch placement and improve the efficacy of our attacks, we utilize the cross-attention mechanism, which encapsulates intermodal interactions by generating attention maps to guide strategic patch placements.
Comprehensive experiments conducted in a white-box setting for image-to-text scenarios reveal that our proposed method significantly outperforms existing techniques, achieving a 100\% attack success rate. Additionally, it demonstrates commendable performance in transfer tasks involving text-to-image configurations.


\end{abstract}

\section{Introduction}
\label{sec:intr}
The visual-language pre-training (VLP) models in the multimodal domain have garnered considerable attention due to their robust performance across a range of visual-language tasks. Currently, VLP models are primarily applied in three downstream tasks: 1) Visual-Language Retrieval \cite{chen2023vlp}: This task involves matching visual data with corresponding textual data. It consists of two sub-tasks: image-to-text retrieval (TR), which retrieves textual descriptions for given images, and text-to-image retrieval (IR), which finds matching images for specific texts. 2) visual entailment (VE)\cite{xie2019visual}: This task uses images and text as premises and hypotheses to predict whether their relationship is entailment, neutral, or contradiction. 3) visual grounding (VG)\cite{hong2019learning}: This task aims to localize object regions in images corresponding to specific textual descriptions. As deep networks are susceptible to error patterns~\cite{liang2023badclip,liu2023pre,liu2023does,liang2024poisoned,liang2024vl,zhang2024towards,zhu2024breaking,liang2024revisiting}, \textit{i.e.}, adversarial perturbations~\cite{liang2021generate,liang2020efficient,wei2018transferable,liang2022parallel,liang2022large,wang2023diversifying,liu2023x,he2023generating,liu2023improving,he2023sa,muxue2023adversarial,lou2024hide,kong2024environmental,ma2021poisoning,ma2022tale,ma2024sequential}, the security of VLP models has also come under scrutiny. Recent studies indicate that VLP models remain vulnerable to adversarial examples\cite{zhang2022towards}. Research into adversarial attacks on VLP models can further enhance their robustness and security~\cite{sun2023improving,liu2023exploring,liang2023exploring,zhang2024lanevil,wang2022adaptive,liang2024unlearning}.

When dealing with multimodal models, attackers can individually target different modalities to reduce the accuracy of downstream tasks. Co-Attack pioneered collaborative attacks by innovatively considering the attack relationships between modalities. Recent research has started to focus on the adversarial transferability of VLP models. However, these attacks are limited in adversarial perturbations and cannot be applied in the physical domain. Typically, attackers use adversarial patch training methods to achieve physical domain attacks. Additionally, they all attack both images and text simultaneously, where text perturbations are easily detected. For example, Co-Attack transforms the text "a man playing guitar" into "a man playing scoring," which clearly does not meet the requirement of invisibility. Therefore, applying adversarial patch attacks to images enables attacks in the physical domain while preserving textual authenticity. This paper is the first to focus solely on naturalistic adversarial patch attacks against VLP models. As demonstrated in \figref{fig:1}, our method achieves superior attack performance in a white-box setting.

\pgfplotsset{compat=1.18}  

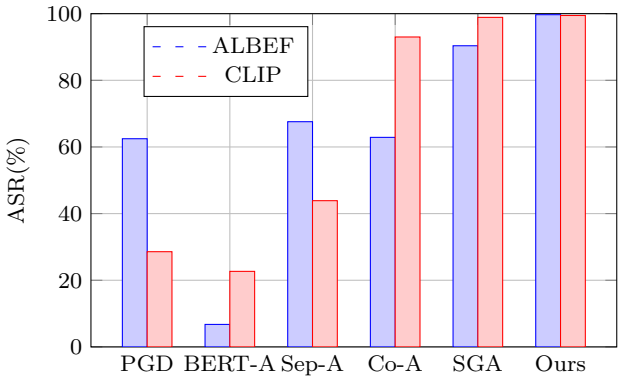
\begin{figure}[h]  
    \centering  
    \begin{tikzpicture}
        \begin{axis}[
            ylabel={ASR(\%)},  
            ymin=0, ymax=100,  
            xtick={0.5,2.5,4.5,6.5,8.5,10.5},  
            xticklabels={PGD, BERT-A, Sep-A, Co-A, SGA, Ours},  
            bar width=0.6,  
            width=8.5cm,
            height=6cm,
            legend style={at={(0.1,0.75)}, anchor=south west},  
            grid=major,  
            ]
            \addplot[
                ybar=2,
                bar width=0.6,
                color=blue,
                fill=blue!20
                ]
                 coordinates {(0.2,62.47) (2.2,6.75) (4.2,67.58) (6.2,62.88) (8.2,90.37) (10.2,99.69)};
            \addlegendentry{ALBEF}

            \addplot[
                ybar=2,
                bar width=0.6,
                color=red,
                fill = red! 20
                ]
                coordinates {(0.8,28.57) (2.8,22.67) (4.8,43.88) (6.8,93) (8.8,98.89) (10.8,99.45)};
            \addlegendentry{CLIP}
        \end{axis}
    \end{tikzpicture}
      \caption{Comparison of attack success rates (ASR) of different attacks in the white box settings (ALBEF\cite{li2021align} and CLIP\cite{radford2021learning}) on image-text retrieval. Starting from left to right as image-only PGD attack\cite{mkadry2017towards}, text-only BERT-Attack, the combined separate unimodal attack (Sep-Attack), Collaborative Attack (Co-Attack\cite{zhang2022towards}), Set-level Guidance Attack (SGA\cite{lu2023set}) and our method.}
  \label{fig:1}       
\end{figure}

However, applying single-modal attacks to multimodal models is challenging and requires leveraging information from the other modality. Co-Attack modified the loss function based on previous work to achieve bimodal collaborative attacks, while SGA considered the similarity between set-level text and images, but neither considered the structure within the victim model. VLP models often employ attention mechanisms for modality interaction internally, which attackers should exploit to construct attacks. Conventional adversarial patch attacks suffer from naturalness defects, inspiring us to use diffusion models to guide adversarial patch generation and create natural adversarial patches. \tabref{tab:1} illustrates the characteristics of different multimodal attack methods, highlighting significant advantages in various aspects of our approach.

\begin{table}[htb]
  \centering
	\renewcommand{\arraystretch}{1.1}
	\setlength{\tabcolsep}{0.7mm}
  \begin{tabular}{lllll} \hline 
  & Image-Attack& Text-Attack & Natural & Physical  \\ \hline
   PGD  &  \checkmark &        &     &  \\
    BERT-Attack  &   &  \checkmark  &  \\
    Sep-Attack &  \checkmark  &   \checkmark    &  \\
    Co-Attack  &  \checkmark &  \checkmark   &  \\
    SGA  &  \checkmark &  \checkmark   &   &  \\
    Ours  &  \checkmark &        & \checkmark  & \checkmark \\ \hline
   
  \end{tabular}
  \caption{Comparison of characteristics of different attack methods.}
  \label{tab:1}       
  \vspace{-2em}
\end{table}

We conducted experiments on two mature multimodal datasets, Flickr30K\cite{plummer2015flickr30k} and MSCOCO\cite{lin2014microsoft}, to evaluate the performance of our proposed method in the task of image-text retrieval. The experimental results demonstrate that our method achieves a balance between attack effectiveness and naturalness across multiple VLP models. Moreover, it exhibits excellent transfer performance, benefiting from cross-attention mechanisms that integrate common features across modalities. This allows adversarial patches to achieve strong attack performance without requiring large perturbations (maintaining a distribution similar to real images). We summarize our contributions as follows:

1)To the best of our knowledge, we are the first to explore the security of VLP models through adversarial patches.
2)We introduce a novel diffusion-based framework to generate more natural adversarial patches against VLP models.
3)We determine the location of adversarial patches by cross-modal guidance. Extensive ablation experiments demonstrate the effectiveness of this approach.

\section{Releated Work}
\subsection{Adversarial Patch}
Adversarial patch attacks can be mainly divided into iterative-based and generative-based methods.

\textbf{Iterative-based methods}. 
Brown et al.\cite{brown2017adversarial} presents a method to create universal, robust, targeted adversarial image patches in the real world.
DPatch~\cite{liu2018dpatch} generates a black-box adversarial patch attack for mainstream object detectors by randomly sampling adversarial patch locations and simultaneously attacking the regression module and classification module of the detection head.
Based on DPatch, Lee et al.~\cite{lee2019physical} use the PGD~\cite{madry2017towards} optimization method as a prototype to generate a more aggressive attack method by randomly sampling patch angle and scale changes.
Pavlitskaya et al.~\cite{pavlitskaya2022suppress} also reveal that the adversarial patch scale is proportional to the attack success rate.
Thys et al.~\cite{thys2019fooling} introduce an adversarial patch attack designed to attack person detection in the physical domain.
Saha et al.~\cite{saha2020role} analyze the attack principle of adversarial patches that do not overlap with the target and propose to use contextual reasoning to fool the detector.
To reduce patch visibility and enhance the attacking ability of the adversarial patch, a large number of works have made a lot of efforts to generate various patches. Specifically, they include adversarial semantic contours that target instance boundaries~\cite{zhang2021adversarial}, adversarial patch groups at multiple locations~\cite{zhao2020object, zhu2021you}, patch-based sparse adversarial attacks~\cite{bao2020sparse}, diffuse patches of asteroid-shaped or grid-shape~\cite{wu2010dpattack}, deformable patch~\cite{chen2022shape} and the translucent patch~\cite{zolfi2021translucent}.

\textbf{Generative-based methods}.
Attacking ability is not the only goal we pursue. The mainstream method to generate an adversarial patch currently is iterative-based which can optimize for the patch to attack the detector without any constraints, while the patch will be generated in an unpredictable direction. 
To address this problem, generative-based methods are considered to trade off Naturalness for attack performance. 
PS-GAN~\cite{liu2022efficient} proposes a perceptual-sensitive generative adversarial network that treats the patch generation as a patch-to-patch translation via an adversarial process, feeding any types of seed patch and outputting the similar adversarial patch with high perceptual correlation with the attacked image. Pavlitskaya et al.~\cite{pavlitskaya2022feasibility} have shown that using a pre-trained GAN helps to gain realistic-looking patches while preserving the performance similar to conventional adversarial patches. 
Hu et al.~\cite{hu2021naturalistic} present a technique for creating physical adversarial patches for object detectors by utilizing the image manifold learned by a pre-trained GAN on real-world images.
There is some work~\cite{wang2023semantic,
dai2023advdiff,
liu2023instruct2attack} beginning to use diffusion models in adversarial attacks.
Diff-PGD~\cite{xue2024diffusion} utilizes a diffusion model-guided gradient to ensure that adversarial samples stay within the vicinity of the original data distribution while preserving their adversarial potency.

\subsection{VLP Model}
Visual language pre-training (VLP) models leverage deep learning techniques to pre-train models on large-scale data, integrating visual and language modalities. As research has progressed, several representative models have emerged.

Early VLP models explored integrating visual and language information into a unified framework to enhance performance across multimodal tasks. With the rise of pre-training methods, a series of new models have been developed. For instance, CLIP\cite{radford2021learning}, developed by OpenAI, achieves strong correlations between images and text through contrastive learning, demonstrating excellent performance across various visual language tasks. Another notable model is BLIP\cite{li2022blip}, which introduces logical reasoning tasks to enhance performance in visual and textual reasoning tasks. Recent advancements include the ALBEF\cite{li2021align} model, which employs enhanced multimodal data augmentation techniques to improve generalization on diverse datasets. Moreover, the TCL\cite{yang2022vision} model proposed by Google focuses on mapping textual descriptions into visual feature spaces, facilitating tasks such as text-to-image retrieval and generation. Additionally, models like ViLBERT\cite{lu2019vilbert} and UNITER\cite{chen2020uniter} have shown outstanding performance in tasks such as image captioning and visual question answering. Together, these models represent the forefront of advancements in integrating and leveraging visual and language information within the VLP domain.

Several studies are currently investigating adversarial attacks on VLP models. Co-Attack\cite{zhang2022towards} posits that standard adversarial attacks are designed for classification tasks involving only a single modality. VLP models engage multiple modalities and often deal with numerous non-classification tasks, such as image-text cross-modal retrieval. Hence, directly adopting standard adversarial attack methods is impractical. Moreover, to target the embedded representations of VLP models, adversarial perturbations across different modalities should be considered collaboratively rather than independently. Our proposed method demonstrates that, in addition to multimodal collaborative attacks, information from other modalities can also be utilized for single-modal attacks. SGA\cite{lu2023set} introduces an ensemble-level guided attack method. This approach extends single image-text pairs to ensemble-level image-text pairs and generates adversarial examples with strong transferability, supervised by cross-modal data.TMM\cite{wang2024transferable} proposes the attention-directed feature perturbation to disturb the modality-consistency features in critical attention
regions.

\section{Preliminaries}
\subsection{Threat Model}
The attacker aims to find a patch $\textbf{\emph{P}}$, which usually follows a square-sized setting where $\textbf{\emph{P}}$ $\in$ $\mathbb{R}^{s\times s\times 3}$ and s accounts for the patch size, into the visual inputs of the VLP models, leading to incorrect outputs in downstream tasks that rely on these pre-training models. Given a benign
image-text pair $d = \{\textbf{\emph{d}}_\mathrm{v}, d_\mathrm{t}\}$, a VLP model can
encode this input into a fused embedding $e$ and $\textbf{\emph{P}}$ is designed to mislead the surrogate
model $\mathcal{F} $ into producing an incorrect embedding:
\begin{equation}
    \mathcal{F}((1-\textbf{\emph{m}})\odot \textbf{\emph{d}}_\mathrm{v}+\textbf{\emph{m}}\odot \textbf{\emph{P}},d_\mathrm{t})  \neq \mathrm{e},
\end{equation}
where $\textbf{\emph{m}}$ denotes a constructed binary mask that is 1 at the placement position of the adversarial patch and 0 at the remaining positions, $\odot$ denotes the Hadamard product (element product).

\begin{figure*}[t]
  \includegraphics[width=0.99\linewidth]{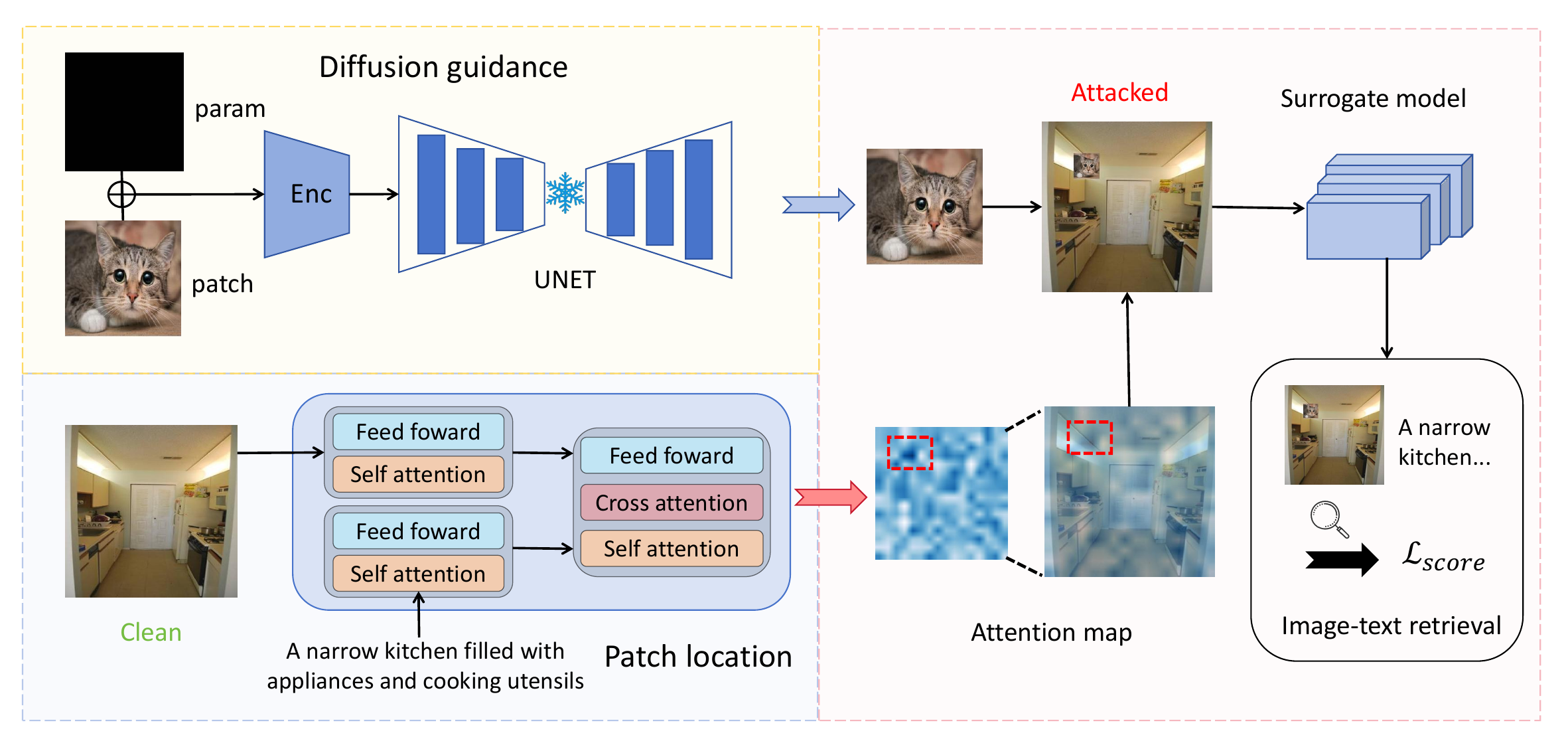}
  \caption{The framework of our proposed Multimodal attack. We employ a dual-guided approach with diffusion and attention mechanisms 
 to balance the attacking ability and the naturalness of adversarial patches.}
  \label{fig:2}       
\end{figure*}

\subsection{Diffusion Models}
We adopt a pre-trained diffusion into our framework. To better understand our work, it is useful to give an overview of Diffusion Models.
Denoising Diffusion Probabilistic Models (DDPM)~\cite{ho2020denoising} is a class of generative models that has gained significant attention in recent years for its ability to produce high-quality samples. 
DDPM consists of two main processes: the forward diffusion process and the denoising process.

The diffusion process is a Markov chain that gradually transforms data points (such as images) into noise. The diffusion process can be represented as:
\begin{equation}
    \textbf{\emph{x}}_\mathrm{t} = \sqrt{\alpha_\mathrm{t}} \textbf{\emph{x}}_\mathrm{t-1} + \sqrt{1 - \alpha_\mathrm{t}} \epsilon_\mathrm{t}, \quad t = 1, 2, \ldots, {\rm T}
\end{equation}
where $x_\mathrm{t}$ is the image at step $t$, $\alpha_\mathrm{t}$ is the diffusion coefficient (which typically decreases with increasing $t$), $\epsilon_\mathrm{t}$ is noise drawn from a standard normal distribution, and T is the number of diffusion steps.

The denoising process is the reverse process of the diffusion process, aiming to recover the original data from the noise. In the Diffusion Model, the denoising process is usually implemented by a conditional neural network (such as U-Net) that predicts the original image based on the current noisy image. The denoising process can be represented as:
\begin{equation}
    \textbf{\emph{x}}_\mathrm{t-1} = \frac{1}{\sqrt{\alpha_\mathrm{t}}} \left(\textbf{\emph{x}}_\mathrm{t} - \frac{1 - \alpha_\mathrm{t}}{\sqrt{1 - \bar{\alpha}_\mathrm{t}}} \epsilon_\theta(\textbf{\emph{x}}_\mathrm{t}, \mathrm{t})\right),
\end{equation}
where $\epsilon_\theta$ is the noise predicted by the neural network, and $\bar{\alpha}_\mathrm{t} = \prod_\mathrm{s=1}^\mathrm{t} \alpha_\mathrm{s}$.

\section{The Proposed Method}
\subsection{Motivation}
Our method is proposed based on the following observations. First, the prevailing approach in the multimodal field to launching adversarial attacks on VLP models involves attacking both images and text simultaneously. Co-Attack has demonstrated that it is indeed possible to find such a collaborative attack method that achieves a synergistic effect greater than the sum of its parts. However, attacking an additional modality also increases the likelihood of the attack being detected, while single-modality attacks often fail to achieve the same effectiveness as multimodal attacks, a contradiction that has prompted us to investigate image-only attacks on VLP models. Secondly, perturbation attacks, as a form of digital domain attack, cannot be applied to the physical domain, which poses another limitation. Combining these two points, we have explored transferring textual information to images to conduct adversarial patch attacks on images. However, this also raises another issue: adversarial patch attacks tend not to be as inconspicuous as perturbation attacks. Therefore, inspired by some diffusion work, we are studying diffusion-based methods for generating adversarial patches.
\vspace{-2em}
\subsection{Patch Generation}
To generate adversarial patches, we first have the init patch $\textbf{\emph{P}}_\mathrm{init}$ which is a real image and the pre-trained Diffusion Model (PDM).
We set an image $\textbf{\emph{d}}_\mathrm{p}$ (perturbation), which is the same size as $\textbf{\emph{P}}_\mathrm{init}$, as the training parameter. The generation process of the patch can be formulated as follows and diffusion process is shown in Alg.~\ref{patch}:
\begin{equation}
\textbf{\emph{P}}_\mathrm{final} = {\mathrm{PDM}(\textbf{\emph{P}}_{init}+\textbf{\emph{d}}_\mathrm{p})}.
\end{equation}
We then focus on patch location. Specifically, we utilize cross-attention to fuse the consistency features of images and text to obtain an attention map. After resizing the attention map to match the original image size through linear interpolation, we can identify the critical areas where the model makes its decisions. The patch is then applied to this location, resulting in the modified image. Subsequently, we perform the scoring for the downstream task (image-text retrieval) and calculate the loss function, which is used to adjust the parameters through backpropagation.

The following will provide a more detailed introduction to the method and its function. 
\begin{algorithm}[t]  
  \caption{Patch Generation}  
  \label{patch}  
  \begin{algorithmic}[1]  
    \Require  
         Interaction N, Time step t, Step size s, Adversarial perturbation $\textbf{\emph{d}}_\mathrm{p}$, Learning rate lr
     \Ensure $\textbf{\emph{P}}_\mathrm{final}$
     \For{$n = 1$ to $N$}
     \State $\textbf{\emph{x}}=\sqrt{\alpha_t} (\textbf{\emph{P}}_\mathrm{init}+\textbf{\emph{d}}_\mathrm{p})+\sqrt{1 - \alpha_\mathrm{t}}z$; \ $z\sim \mathcal N(\textbf{0},\textbf{I})$
     \Repeat
     \State $\textbf{\emph{x}}_\mathrm{t} = x$
     \State $\textbf{\emph{x}}_\mathrm{t-s} = \sqrt{\alpha_\mathrm{t-s}} \left( \frac{\textbf{\emph{x}}_\mathrm{t} - \sqrt{1 - \alpha_\mathrm{t}} \cdot \epsilon_{\theta}(\textbf{\emph{x}}_\mathrm{t},t)}{\sqrt{\alpha_\mathrm{t}}} \right) + \sqrt{1 - \alpha_\mathrm{t-s}} \cdot \epsilon_{\theta}(\textbf{\emph{x}}_\mathrm{t},t)$
     \State t = t - s
     \Until t \textless s
     \State $\textbf{\emph{d}}_\mathrm{p} = \textbf{\emph{d}}_\mathrm{p} - lr*\nabla_\mathrm{d} \mathcal{L}_p$ 
     \EndFor
     \State $\textbf{\emph{P}}_\mathrm{final} = \textbf{\emph{x}}_\mathrm{t}$
  \end{algorithmic}  
\end{algorithm}

\subsection{Diffusion Guidance}
Currently, the majority of adversarial patch methods directly optimize the adversarial patch itself, but this approach can cause significant changes to the original image to achieve good attack effects, which poses a great challenge to the naturalness of the adversarial patch. In contrast, since there are no hidden layers in the network, the model parameters can be set to a tensor $\textbf{\emph{d}}_\mathrm{p}$ with the same size as $\textbf{\emph{P}}_\mathrm{init}$ and a value of zero. Compared to directly optimizing the patch, adding adversarial perturbations has many advantages. Firstly, the perturbation can be seen as noise in the original image, which better matches the denoising process of diffusion model, and makes it easier to find constrained optimal solutions. Secondly, this method involves fewer changes to the original image and it can preserve the information of the original image. From a macroscopic perspective, similar to PGD, it is like adding adversarial perturbations to $\textbf{\emph{P}}_\mathrm{init}$.

We exploit the $l_{\infty}$ norm to constrain d, and the formula for updating $P_{init}$ in each iteration is as follows:
\vspace{-1em}
\begin{equation}
\textbf{\emph{P}}_{init} = {\rm Clip}(\textbf{\emph{P}}_{\mathrm{init}} + \textbf{\emph{d}}_\mathrm{p}).
\end{equation}
Clip is the clipping function defined in Eq. \ref{eq6}.
\begin{equation}
{\rm Clip}(\textbf{\emph{P}}) = \{p_{\mathrm{i}}|p_{\mathrm{i}}\leftarrow \mathrm{min}(\mathrm{max}(p_{\mathrm{i}},\tau),0)\},
\label{eq6}
\end{equation}
where ${p_\mathrm{i}}$ is the i-th element of $\textbf{\emph{P}}$ and $\tau$ is maximum value of ${p_\mathrm{i}}$.

The adoption of diffusion models to guide gradients is primarily aimed at ensuring that adversarial examples remain close to the original data distribution while maintaining their efficacy. This is because existing adversarial attacks, generated using gradient-based techniques in digital and physical scenarios, often diverge significantly from the actual data distribution of natural images, resulting in a lack of naturalness and authenticity. While GAN-based methods can generate realistic images, the adversarial samples are sampled from noise, thus lacking controllability. Therefore, adversarial patch generation based on diffusion models offers significant advantages.

\begin{figure*}[tbp]
    \centering
    
    \begin{tabular}{m{0.07\textwidth}m{0.20\textwidth}m{0.20\textwidth}m{0.20\textwidth}m{0.20\textwidth}}
        \centering clean &
        \includegraphics[width=\linewidth]{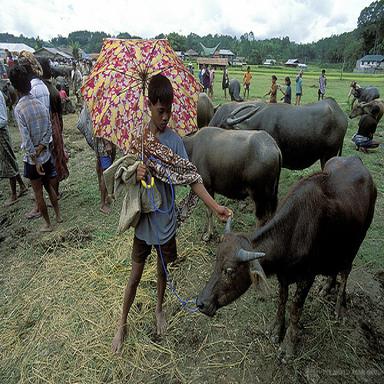} &
        \includegraphics[width=\linewidth]{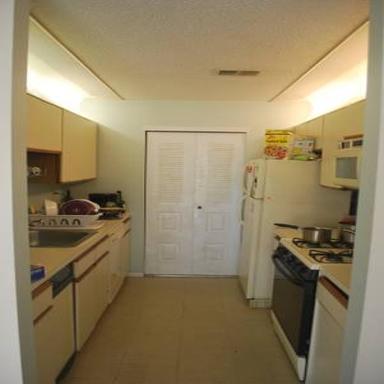} &
        \includegraphics[width=\linewidth]{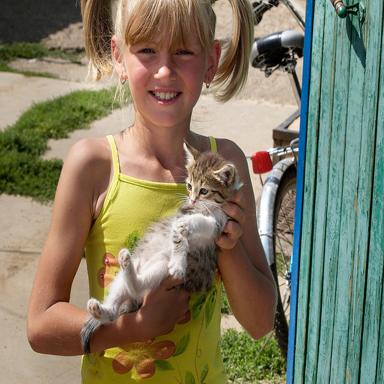} &
        \includegraphics[width=\linewidth]{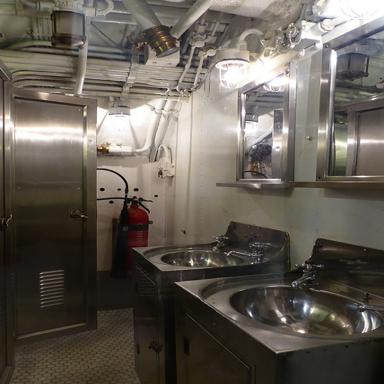} \\
        
        \centering attacked &
        \includegraphics[width=\linewidth]{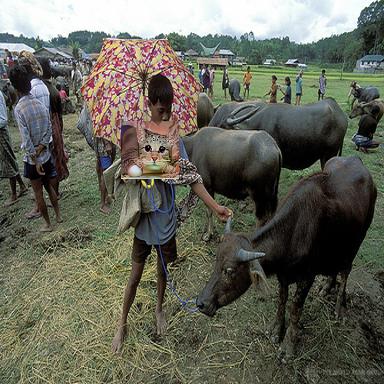} &
        \includegraphics[width=\linewidth]{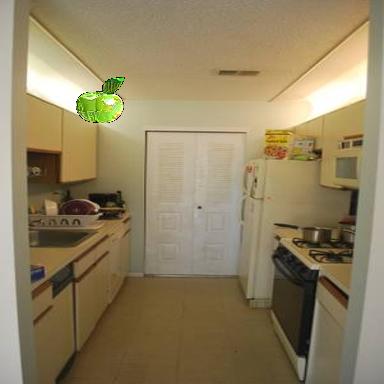} &
        \includegraphics[width=\linewidth]{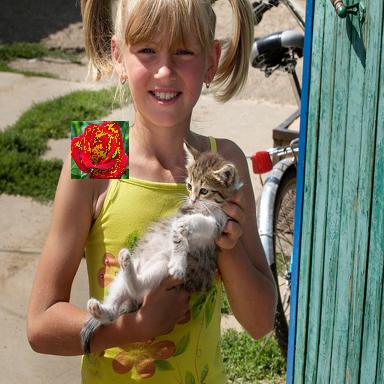} &
        \includegraphics[width=\linewidth]{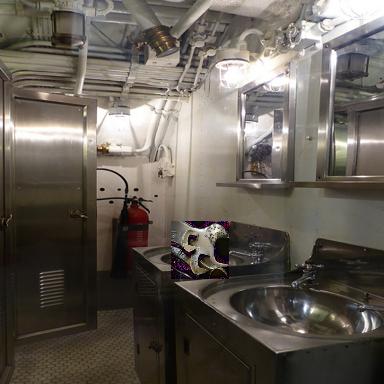} \\
    \end{tabular}
    
    \caption{ The clean images and the attacked images with naturalistic patches. The images shown are from the dataset MSCOCO\cite{lin2014microsoft} }
    \label{fig:3}
    \vspace{-1em}
\end{figure*}

\subsection{Patch Location}
The vast majority of VLP models utilize attention mechanisms to capture the consistency features between image and text. Previous work\cite{gou2023leveraging,lan2019learning} has highlighted that modality consistency features significantly influence the decision-making of multimodal models and are crucial for the success of downstream tasks. Therefore, we believe that in VLP models, the output of the commonly used cross-attention modules designed for cross-modal interaction reflects the text's attention to the image. Some works on region-specific attacks have already demonstrated the importance of attacking specific areas. For adversarial patch attacks, the placement location can affect the success rate and the training process. Placing adversarial patches on vulnerable parts of the image can achieve more with less, meaning attacks can be carried out without significant perturbations. This also helps in maintaining the naturalness of the adversarial patches. Therefore, we use cross-attention to guide the placement of adversarial patches. The attention map $M$ is calculated as follows:
\begin{equation}
    M = \mathrm{softmax}(\frac{QK^{T}}{\sqrt{\mathrm{s}}})V,
\end{equation}
where $Q$,$K$,$V$ denote the feature matrix of different modalities, and $\sqrt{\mathrm{s}}$
denotes the scaling factor for stabilizing the model.
Because the generated attention map M does not match the size of the image, it needs to be resized to the dimensions of the image using bilinear interpolation, with the maximum value inside serving as the central position for the patch.
\vspace{-1em}
\subsection{Loss Function}
Our patch optimization is implemented through the computation of two losses:
\begin{equation}
    \mathcal{L}_\mathrm{p} = \mathcal{L}_\mathrm{score} + \lambda \mathcal{L}_\mathrm{tv}.
\end{equation}
In the third part of the pipeline, the obtained $\textbf{\emph{P}}_\mathrm{final}$ is applied to the clean image $d_\mathrm{v}$ guided by the attention map to produce the attacked image $\hat{d_v}$:
\begin{equation}
    \hat{\textbf{\emph{d}}_\mathrm{v}} = (1-\textbf{\emph{m}})\odot \textbf{\emph{d}}_\mathrm{v}+\textbf{\emph{m}}\odot \textbf{\emph{P}}_\mathrm{final}.
\end{equation}

The image-text pair $d = \{\hat{\textbf{\emph{d}}_\mathrm{v}},d_\mathrm{t}\}$ is input into the VLP model targeted for attack, and the scores for the downstream task are calculated. For a dataset of 1000 images and 5000 texts, each image will receive scores corresponding to 5000 texts. We extract the top $k$ highest scores and divide these 
 scores into two sets, $S_1$ and $S_2$, representing scores of texts that belong or do not belong to the image, respectively. $\mathcal{L}_\mathrm{score}$ is calculated as follows:
\begin{equation}
\label{lscore}
    \mathcal{L}_\mathrm{score} = \mathrm{max}(S_1) - \mathrm{min}(S_2).
\end{equation}

Total variation loss is effective in removing noise while preserving edge information, resulting in smoother and clearer images. Compared to other smoothing techniques, total variation loss better preserves the edges and texture details of images, avoiding excessive blurring.
\begin{equation}
    \mathcal{L}_\mathrm{tv} = \frac{\sqrt{\sum_{i}^{S} \sum_{j}^{S} \left(\textbf{\emph{P}}_\mathrm{i,j} - \textbf{\emph{P}}_\mathrm{i+1,j} \right)^2 + \left( \textbf{\emph{P}}_\mathrm{i,j} - \textbf{\emph{P}}_\mathrm{i,j+1} \right)^2}}{\rm N},
\end{equation}
where N denotes the number of pixels on the given adversarial patch $\textbf{\emph{P}}_\mathrm{final}$.

\vspace{-1em}
\section{Experiment}
\subsection{Implementation details}
\subsubsection{Datasets and VLP Model}
Flickr30K\cite{plummer2015flickr30k} consists
of 31,783 images, each with five corresponding captions.
Similarly, MSCOCO\cite{lin2014microsoft} comprises 123,287 images, and each
image is annotated with around five captions. We adopt the
Karpathy split\cite{karpathy2015deep} for experimental evaluation.
We evaluate two popular VLP models, the
fused VLP and aligned VLP models. For the fused VLP,
we consider ALBEF\cite{li2021align}. ALBEF contains
a 12-layer visual transformer ViT-B/16\cite{dosovitskiy2020image} and two 6-layer
transformers for the image encoder and both the text encoder
and the multimodal encoder, respectively. TCL uses the
same model architecture as ALBEF but with different pre-trained objectives. For the aligned VLP model, we choose to
evaluate CLIP\cite{radford2021learning}. CLIP has two different image encoder
choices, namely, CLIPViT and CLIPCNN, that use ViTB/16 and ResNet-101\cite{he2016deep} as the base architectures for the
image encoder, respectively.

\begin{table*}[t]
\centering
\caption{ Image-text retrieval results of ALBEF and CLIP on MSCOCO dataset and Flickr30K dataset. The reported value is attack success rate(100\%).
}
\setlength{\tabcolsep}{4pt} 
\renewcommand{\arraystretch}{1.2} 
\begin{tabular}{|c|c||c|c|c|c|c|c|c|c|c|c|c|c|}
\hline
 &  & \multicolumn{6}{c|}{MSCOCO (5K test set)} & \multicolumn{6}{c|}{Flickr30K (1K test set)}\\

\cline{3-14} 
Model&  Attack & \multicolumn{3}{c|}{TR} & \multicolumn{3}{c|}{IR} & \multicolumn{3}{c|}{TR} & \multicolumn{3}{c|}{IR}\\

\cline{3-14} &   &R@1 &R@5 & R@10 & R@1 & R@5 & R@10& R@1 & R@5 &R@10 & R@1 &R@5 & R@10\\
 
\hline
 & PGD & 76.7 & 67.49  & 62.47  & 86.3   & 78.49  & 73.94 
&52.45&36.57&30.00&58.65&44.85&38.98       \\
& BERT-Attack   & 24.39 & 10.67 & 6.75 & 36.13  & 23.71 & 18.94     &11.57&1.8&1.1&27.46&14.48&10.98    \\
ALBEF& Sep-Attack  & 82.60  & 73.2  & 67.58  & 89.88 & 82.6  & 78.82      &65.69&47.6&42.1&73.95&59.5&53.7   \\
& Co-Attack & 79.87  & 68.62 & 62.88 & 87.83 & 80.16 & 75.98 
&77.16&64.6&58.37&83.86&74.63&70.13     \\
& SGA  & 96.7 & 92.83 & 90.37  & 96.95  & 93.44 & 91.00   
&97.24&94.09&92.3&97.28&94.27&92.58    \\
& Ours  & 99.90   & 99.69   & 99.69  &  99.90 & 99.49   & 98.97  
&99.78&99.32&99.32&99.78&98.86& 97.72    \\ 
\hline
 & PGD & 54.79&	36.21&	28.57&	66.85&	51.8&	46.02&	70.92&	50.05&	42.28&	78.61&	60.78&	51.5
       \\
& BERT-Attack   & 45.06	&28.62	&22.67&	51.68&	37.12&	31.02&	28.34&	11.73&	6.81&39.08&	24.08&	17.44
    \\
CLIP& Sep-Attack  & 68.52&	52.3&	43.88&	77.94&	66.77	&60.69&	79.75&	63.03&	53.76&	86.79&	75.24&	67.84
   \\
& Co-Attack & 97.98&	94.94&	93.00&	98.80&	96.83&	95.33&	93.25&	84.88&	78.96&	95.68&	90.83&	87.36
     \\
& SGA  & 99.79&	99.37&	98.89&	99.79&	99.37&	98.94&	99.08&	97.25&	95.22&	98.84&	97.53&	96.03
    \\
& Ours & 99.85&	99.73&	99.45&	99.81&	99.23&	98.32&	99.92&	99.68&	99.18&	99.68&	98.26&	97.75
     \\ 
\hline
\end{tabular}
\label{tab:2} 
\end{table*}

\subsubsection{Adversarial Attack Settings and Metrics}To better compare our method with the SoTA method, we mainly use the parameter settings of SGA. We employ PGD with perturbation bound $\epsilon$ = 2/255,
step size $\alpha$ = 0.5/255, and iteration steps T = 10. In our experiment, the diffusion model we adopt is the unconditional diffusion model pre-trained on ImageNet\cite{deng2009imagenet} though we use DDIM to respace the original timesteps for faster inference. In the image-text retrieval task, each image has the top k text scores, where k is set to 15 in the white-box setting. We chose 15\% of the original image as the patch size. In the ablation study, we will explore the impact of different values of k and patch sizes on the attack. We employ the attack success rate (ASR) as the main metric for evaluating the attacking capability of the generated adversarial examples in VLP
downstream tasks. This metric reflects the proportion of adversarial examples that successfully influence the decisions of models. The higher the ASR, the better the attacking
ability. Specifically, we offer ASR values for R@1, R@5, and R@10 in all tables for the
tasks of image-to-text (TR) and text-to-image retrieval (IR), where R@N represents the top N most relevant text/image based on the image/text.

\subsection{Comparisons of SoTA Method}
To rigorously evaluate the superiority of our proposed method within the white-box setting, we conducted comprehensive comparisons with several baseline approaches. These included the image-only PGD attack \cite{madry2017towards}, the text-only BERT-Attack, the combined separate unimodal attack (Sep-Attack), the Collaborative Attack (Co-Attack) \cite{zhang2022towards}, and the Set-level Guidance Attack (SGA) \cite{lu2023set}. These comparisons were performed using the widely recognized test datasets MSCOCO and Flickr30K on both the ALBEF and CLIP models. Representative samples of clean and adversarial images are illustrated in \figref{fig:3}.

Our method, guided by the cross-attention and diffusion model, successfully maintains the adversarial patch close to the real image distribution, thereby striking an optimal balance between naturalness and attack efficacy. To further validate the robustness of our adversarial examples, we introduced noise to the generated adversarial samples. During training, the parameter 
K was set to 15, and the attack iterations were continued until the loss was minimized. This methodology ensures that, for an image with only five corresponding texts, the attack success rate in the text retrieval (TR) task for Recall@10 (R@10) reaches 100\%.

As demonstrated in \tabref{tab:2}, our method consistently outperforms other techniques in the white-box setting. On average, with the ALBEF model, our approach surpasses the state-of-the-art methods by \textbf{6.46\%} and \textbf{4.93\%} in the TR task on the MSCOCO and Flickr30K datasets, respectively. When applied to the image retrieval (IR) task, we achieve improvements of \textbf{5.65\%} and \textbf{4.07\%}. Notably, similar performance enhancements were observed with the CLIP model.

An important aspect of our approach is the utilization of cross-attention to integrate information from both images and texts, thereby obtaining the text's attention on the image. It is noteworthy that, despite the CLIP model not performing explicit image-text fusion operations, our method remains effective, demonstrating its versatility and robustness across different model architectures.

\subsubsection{Discussion of Naturalness}
\begin{figure}[htbp]
\vspace{-1em}
    \centering
    \begin{tabular}{cc}

\includegraphics[width=0.22\textwidth]{att3.jpg} & \includegraphics[width=0.22\textwidth]{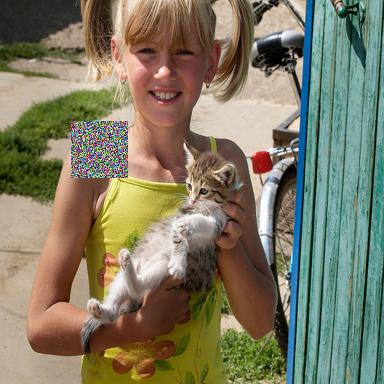} \\

\parbox[t]{0.22\textwidth}{\centering Natural (flower)} & 
\parbox[t]{0.22\textwidth}{\centering Unnatural (noise)} \\

\end{tabular}
    
    \caption{ Comparison of adversarial patches with and without naturalness. The clean images and the attacked images with naturalistic patches. The images shown are from the dataset MSCOCO\cite{lin2014microsoft} }
    \label{natural1}
    \vspace{-1em}
\end{figure}

Previous work has scarcely discussed the naturalness of adversarial patches and lacks related definitions and evaluation methods. We consider that natural adversarial patches should be inconspicuous within adversarial examples. Our approach enables the selection of the most suitable adversarial patches for specific images.\figref{natural1} compares natural adversarial patches with unnatural ones. We chose a rose as the adversarial patch and placed it on the right shoulder of the girl, making it easily mistaken for a part of the clothing decoration. It is noteworthy that through extensive experiments, we found that high-attention areas are often not the most prominent parts, such as the face, which greatly aids in enhancing naturalness.

Naturalness contribute to both inconspicuousness and the final performance. We propose Segment and Complete (SAC)\cite{liu2022segment} to evaluate the robustness of our naturalistic adversarial patches against defender. Our experiments demonstrate that the adversarial patches we generate cannot be detected by defender (detection success rate of patches is 0$\%$).

\subsection{Ablation Study}
In this section, we further investigate the critical factors that influence our proposed method.

\subsubsection{Top K}
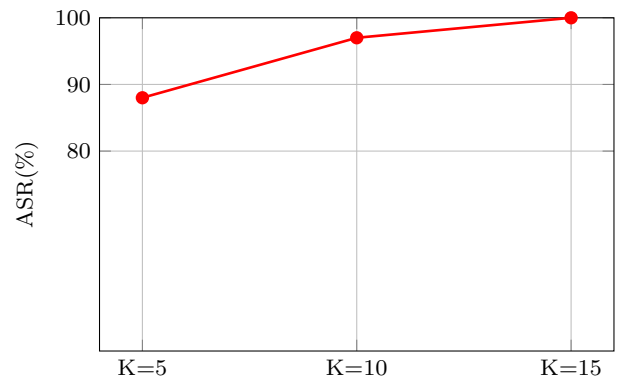
\begin{figure}[ht]
    \centering
    \caption{The mean of ASR on ALBEF and CLIP under different K settings.  }
  \label{fig:4}       
    \begin{tikzpicture}
        \begin{axis}[
            width=0.48\textwidth,
            height=6cm,
            ylabel={ASR(\%)},
            ylabel near ticks, 
            ymin=50, ymax=100,
            xtick={0,1,2},
            xticklabels={K=5, K=10, K=15},
            ytick={80,90,100}, 
            grid=major
        ]
        \addplot[
            color=red,
            mark=*,
            line width=1pt
            ] coordinates {
            (0, 88)
            (1, 97)
            (2, 100)
            };

        \end{axis}
    \end{tikzpicture}
\end{figure}

The choice of k is important for the training process of generating adversarial patches. It is evident that as long as k is greater than 15, white-box attacks can be successful. \tabref{tab:2} also shows that the generated adversarial samples exhibit a certain degree of robustness and perform well in transfer tasks. However, during the experiments, we found that increasing k leads to a higher number of attack iterations, causing the generated adversarial patches to lose their naturalness. Therefore, we experimented with different values of k to attack ALBEF and CLIP, exploring a more suitable choice of k.\figref{fig:4} shows the change in ASR when K takes different values under the condition that the patch size is fixed at 15\%. As K increases from 5 to 15, the ASR increases from 88\% to 100\%. It can be seen that our method can still achieve an ASR of 88\% even when maintaining a very high level of naturalness (K=5).
\vspace{-1em}
\vspace{-2em}
\subsubsection{Patch Location}
\begin{figure}[htbp]
\vspace{-1em}
    \centering
    \begin{tabular}{cc}

\includegraphics[width=0.22\textwidth]{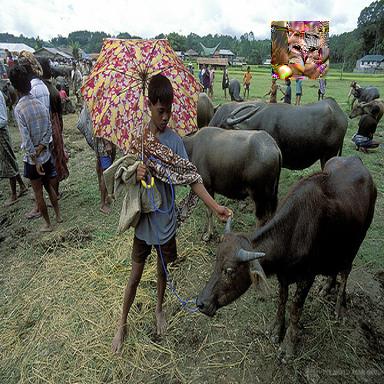} & \includegraphics[width=0.22\textwidth]{att1.jpg} \\

\parbox[t]{0.22\textwidth}{\centering Random Location} & 
\parbox[t]{0.22\textwidth}{\centering Designed Location} \\

\end{tabular}
    
    \caption{ The adversarial examples under different location strategies.The clean images and the attacked images with naturalistic patches. The images shown are from the dataset MSCOCO\cite{lin2014microsoft} }
    \label{fig:5}
\end{figure}

\begin{figure}[ht]
    \centering
    \caption{The mean of attack iteration on ALBEF and CLIP under different location strategies. }
  \label{fig:6}       
    \begin{tikzpicture}
        \begin{axis}[
            width=0.48\textwidth,
            height=6cm,
            ymin=0, ymax=100,
            ylabel={Attack iteration},
            ylabel near ticks, 
            xtick={0,1},
            xticklabels={Random location, Designed location},
            ytick={20,30,40,50,60,70,80,90,100}, 
            grid=major
        ]
        \addplot[
            color=red,
            mark=*,
            line width=1pt
            ] coordinates {
            (0, 35)
            (1, 22)
            };
        \addlegendentry{Random optimization}
                \addplot[
            color=blue,
            mark=*,
            line width=1pt
            ] coordinates {
            (0, 78)
            (1, 38)
            };
            \addlegendentry{Diffusion-base optimization}
        \end{axis}
    \end{tikzpicture}
    \vspace{-2em}
\end{figure}
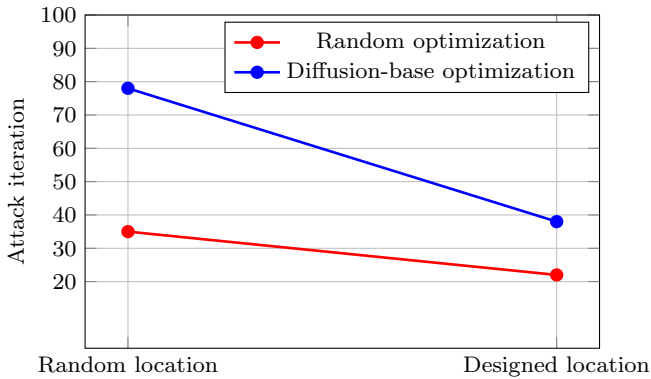

We conducted ablation experiments on the patch location. \figref{fig:5} and \figref{fig:6} illustrate the changes in adversarial patches and the number of attack iterations under different localization strategies. 

 We fixed the patch size at 15\% of the image and set K to 15 to compare the effect of having or not having patch localization on generating adversarial patches. It is evident that, compared to random localization, attention-guided localization can effectively identify suitable attack regions, completing the attack with fewer iterations. This results in reducing time (93s to 45s) for generating an adversarial example and increased naturalness of the adversarial patches.
\vspace{-1em}
\subsubsection{Patch Size}
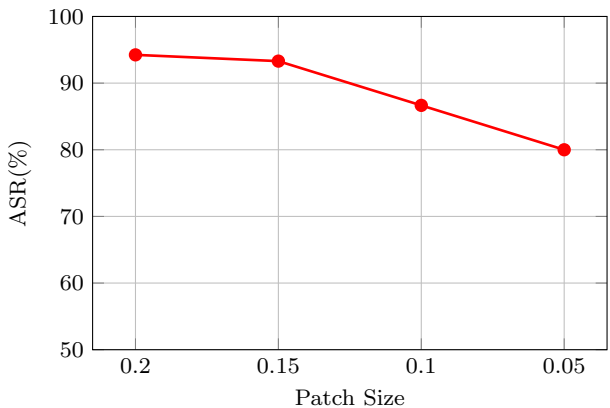
\begin{figure}[H]
    \centering
    \caption{ The attack success rates of different
patch sizes.}
    \begin{tikzpicture}
        \begin{axis}[
            width=0.48\textwidth,
            height=6cm,
            xlabel={Patch Size},
            ylabel={ASR(\%)},
            ylabel near ticks, 
            ymin=50, ymax=100,
            xtick={0,1,2,3},
            xticklabels={0.2, 0.15, 0.1, 0.05},
            ytick={50, 60, 70, 80,90,100}, 
            grid=major
        ]
        \addplot[
            color=red,
            mark=*,
            line width=1pt
            ] coordinates {
            (0, 94.24)
            (1, 93.3)
            (2, 86.66)
            (3, 80)
            };

        \end{axis}
    \end{tikzpicture}
    \label{fig:size}
\end{figure}

We define patch size as the ratio of the length (or width) of the patch to the length (or width) of the image. We set K to 10 to compare the attack success rates of different patch sizes under a white-box setting. To prevent the adversarial patches from degrading into noisy images during training, we set the maximum number of attack iterations to 300. \figref{fig:size} shows the changes in attack success rates and adversarial patches as the patch size varies from 0.2 to 0.05. It is evident that larger adversarial patches achieve more effective attacks and result in more natural-looking adversarial patches.
\vspace{-1em}

\section{Conclusion}
This paper is the first to consider using adversarial patch attacks exclusively on VLP models. By employing a dual-guided approach with diffusion and attention mechanisms, we control the optimization direction and determine the placement of the patches. We propose a framework for generating natural patches that attack image-text retrieval tasks of VLP models while keeping the text unchanged. Our experiments demonstrate the superiority and feasibility of the method.

\textbf{Limitation.}While our method exhibits excellent performance in white-box settings and transfer tasks, experiments reveal a lack of model transferability. We believe this is due to the insufficient utilization of the consistency features between images and text during the attack. The natural adversarial patch attacks makes it more challenging to leverage text attention compared to digital domain perturbation attacks. Additionally, the robustness of physical attacks requires further improvement.
\bibliographystyle{unsrt}
\bibliography{reference}

\newpage
\section*{List of abbreviations}
VLP: visual language pre-training; ASR: attack success rates;
TR: image-to-text retrieval; IR: text-to-image retrieval; VE: visual entailment; VG: visual grounding

\section*{Declarations}
\begin{enumerate}
\item {\bf Availability of data and material}
    \item[] The datasets generated during and/or analyzed during the current study are available from the corresponding author on reasonable request.
\item {\bf Competing Interests }
    \item[] The authors have no competing interests to declare that are relevant to the content of this article.
\item {\bf Author Contributions  }
    \item[] To the best of our knowledge, we are the first to explore the security of VLP models through adversarial patches.We introduce a novel diffusion-based framework to generate more natural adversarial patches against VLP models.
We determine the location of adversarial patches by cross-modal guidance. Extensive ablation experiments demonstrate the effectiveness of this approach.
\item {\bf Funding }
    \item[] This work was supported by the Shenzhen Campus of Sun Yat-sen University.
\item {\bf Acknowledgements }
    \item[] Not applicable. 

\end{enumerate}

\end{document}